\documentclass[10pt,twocolumn,letterpaper]{article}

\usepackage{times}
\usepackage{epsfig}
\usepackage{graphicx}
\usepackage{amsmath}
\usepackage{amssymb}

\usepackage{booktabs}
\usepackage{multirow}
\usepackage{colortbl}
\usepackage[table,xcdraw]{xcolor}

\usepackage[pagebackref=true,breaklinks=true,colorlinks=true,linkcolor=blue,citecolor=blue,urlcolor=blue]{hyperref}

\usepackage{algorithm}
\usepackage{algpseudocode}

\usepackage{caption}
\usepackage{subcaption}
\usepackage[square,numbers]{natbib}  % If you go with numbered references

\usepackage{mathtools}

\usepackage[margin=1in]{geometry}

\begin{document}

%%%%%%%%% TITLE
\title{SegVec3D: A Method for Vector Embedding of 3D Objects Oriented Towards Robot manipulation}

\author{
Zhihan Kang \quad Boyu Wang \\
Northwestern Polytechnical University, Xi'an, China \\
}
\maketitle

%%%%%%%%% ABSTRACT
\begin{abstract}
3D point clouds have become a fundamental data format in robotic perception and manipulation tasks, widely applied in semantic segmentation, instance recognition, and human–robot interaction. However, due to their inherent sparsity, disorder, and lack of structure, instance-level semantic understanding of point clouds remains challenging – particularly under conditions of limited supervision and cross-modal semantic ambiguity. To address these issues, we propose \textbf{SegVec3D}, a novel framework integrating attention mechanisms, embedding learning, and cross-modal alignment techniques for 3D point cloud instance segmentation. The proposed approach first builds a hierarchical instance feature extractor based on spatial adjacency and attention computation, enhancing the model’s ability to capture fine-grained geometric structures. It then introduces a high-dimensional embedding space, enabling \emph{unsupervised} instance segmentation through a contrastive-learning-based clustering mechanism. Furthermore, a shared cross-modal semantic space is constructed to align 3D data with natural language descriptions, allowing zero-shot understanding and retrieval of 3D objects given text queries. The model is ultimately deployed and validated in realistic scenarios, demonstrating strong generalizability and engineering feasibility. While recent methods like Mask3D~[40] and ULIP~[10][11] have advanced 3D segmentation and vision-language pre-training respectively, our approach uniquely integrates these domains by enabling instance segmentation with minimal labeling and directly aligning point clouds with language. Experimental evaluations confirm that the proposed method achieves high semantic discriminability, robust multi-modal alignment, and practical deployability. It supports weakly-supervised or unsupervised 3D instance understanding, providing a promising foundation for future multi-modal cognitive robotic systems.
\end{abstract}

%%%%%%%%% INTRODUCTION
\section{Introduction}
Robots operating in the real world require an effective understanding of the 3D environment, yet they lack the innate 3D perception abilities of humans. In complex manipulation tasks such as grasping objects on a table or navigating around obstacles, precise perception and semantic understanding of 3D objects are critical. A common representation of 3D environments is the \textit{point cloud} – a set of points sampled from surfaces in 3D space – typically produced by sensors like LiDAR or depth cameras. Point clouds preserve rich geometric structure, offering an irreplaceable spatial perspective for robotics tasks. However, point cloud data are characteristically sparse, unordered, and lack explicit topological connections, making them far more difficult for machines to process than 2D images. These challenges are exacerbated under limited annotation, where achieving high-precision instance segmentation remains an open problem.

Recent research has also emphasized the importance of \textit{cross-modal semantic understanding} in robotics. Beyond purely visual perception, a robot should interpret human instructions (e.g., “bring me the wooden chair”) and ground them in the 3D scene. This demands a shared semantic embedding space bridging language, vision, and 3D data. Pioneering vision-language models like CLIP~[5] have demonstrated powerful zero-shot recognition by aligning image and text embeddings. Extensions of such models to 3D, including CLIP2Point~[21] and ULIP~[10], have started to enable language-driven understanding of 3D data. However, achieving effective \textit{3D point cloud–language alignment} for open-world robotic tasks is still in its infancy.

Traditional methods for 3D object recognition relied on heuristic feature engineering and geometric modeling (e.g., region growing, normal clustering), which suffered from low efficiency and poor generalization. The advent of deep learning, especially graph neural networks (GNNs) and attention mechanisms, has opened new avenues for learning semantic representations from unstructured 3D data. For instance, dynamic graph convolutions and point-based networks can capture local geometric patterns in point clouds~[2][8], and transformer-style attention can adaptively weight important points~[9][19]. These developments have substantially improved point cloud segmentation performance. Most recently, transformer-based architectures such as Mask3D~[40] have achieved state-of-the-art results in 3D instance segmentation by introducing learnable object queries and mask prediction in 3D scenes.

Despite this progress, there remain notable gaps. Current 3D instance segmentation models generally require dense annotations and do not incorporate cross-modal information, while multi-modal models like ULIP~[10][11] excel at aligning modalities but do not perform fine-grained 3D segmentation. In contrast, our work focuses on combining \textbf{instance segmentation} with \textbf{cross-modal embedding} under weak or no supervision. We propose an end-to-end system that can segment 3D point cloud instances without heavy labeling and simultaneously align the segmented instances with natural language embeddings.

In summary, the main contributions of this work include:
\begin{itemize}
  \item We propose a novel perspective for 3D object modeling based on spatial density and adjacency, establishing that \emph{relative spatial relationships} take precedence over absolute coordinates for point cloud structure modeling. This forms a foundation for subsequent instance segmentation and embedding.
  \item We design an \textbf{attention-based instance segmentation network} for point clouds. This network leverages a local feature graph and learnable attention to aggregate point features, enabling adaptive focus on relevant neighbors and improving segmentation of complex geometries.
  \item We introduce a \textbf{discriminative embedding space} for 3D instances using a contrastive learning strategy. By projecting points into a high-dimensional vector space and applying a pull-push loss, the network can cluster points belonging to the same object instance without explicit labels.
  \item We develop a \textbf{cross-modal semantic alignment mechanism} that bridges 3D point clouds and natural language. Drawing inspiration from CLIP, we jointly embed point cloud data and textual descriptions into a shared space, enabling zero-shot identification of 3D objects from language queries.
  \item We implement and deploy the proposed SegVec3D system in real-world settings and perform qualitative evaluations on representative scenarios. We demonstrate the system’s robustness and discuss its performance, limitations, and future improvements for multi-modal robotic perception.
\end{itemize}

\noindent The rest of the paper is organized as follows: Section 2 reviews related work on point cloud segmentation and cross-modal learning. Section 3 details the attention-based 3D instance segmentation network design. Section 4 describes the cross-modal semantic embedding approach. Section 5 presents experimental results and analyses. Finally, Section 6 concludes the paper with future outlook.

%%%%%%%%% RELATED WORK
\section{Related Work}
\paragraph{3D Instance Segmentation.} Early 3D segmentation methods relied on handcrafted features and geometric heuristics, such as region growing or surface normal clustering, to partition point clouds into segments~[13]. These classical approaches could segment simple scenes but struggled with complex, real-world data and had limited robustness to noise. The introduction of deep learning revolutionized 3D instance segmentation. PointNet~[2] was a seminal work that processed points directly with multilayer perceptrons and symmetric functions, establishing a foundation for learning point-wise features. PointNet++~[14] extended this by hierarchically aggregating local neighborhoods, capturing multi-scale context. Subsequent methods explored efficient convolutions and data structures for point clouds, such as KPConv~[15] and sparse voxel networks, to scale to large scenes. Graph-based neural networks like DGCNN~[8] modeled point clouds as graphs, applying edge convolutions to learn local geometric relationships. Recent top-performing models often combine point-based and voxel-based techniques for accuracy and efficiency; for example, SoftGroup~[17] uses a two-stage approach (semantic segmentation then clustering) with a group merging strategy to produce instances. More relevantly, transformer architectures have emerged: \textbf{Mask3D}~[40] is a transformer-decoder model that represents each object instance with a learned query vector (analogous to DETR for images). Mask3D attends to point features and directly predicts point masks for all instances in parallel, achieving state-of-the-art results on indoor datasets. Unlike Mask3D, which is fully-supervised and relies on mask annotations, our approach learns an embedding for each point and separates instances via unsupervised clustering in feature space, making it applicable even with sparse or no ground-truth labels.

\paragraph{Graph Modeling and Attention in 3D.} Graph neural networks (GNNs) have provided a powerful framework for 3D data by treating points as nodes in a graph with edges defined by spatial proximity~[8][19]. This representation naturally captures the point cloud’s unstructured nature. The Dynamic Graph CNN (DGCNN)~[8] introduced the idea of dynamically updating the neighbor graph in each layer based on learned feature distances, greatly enhancing local feature learning. Graph Attention Networks (GAT)~[19] brought attention mechanisms to graph nodes, enabling each point to learn weights for its neighbors rather than treating all neighbors equally. In the 3D domain, applying attention has improved the flexibility of feature aggregation. For example, Point Transformer~[9] applies self-attention to point sets, allowing the network to concentrate on the most relevant points for a given task. Our method also employs attention: each point dynamically attends to its $k$ nearest neighbors, as described in Section 3.1.2. Through a trainable attention weight $\alpha_{ij}$, the network selectively aggregates information from neighbors (instead of a uniform average), focusing on those that contribute most to the point’s semantic context. This approach improves robustness to variations in local geometry and density. We further stack multiple attention layers hierarchically to capture long-range dependencies in the point cloud, something traditional graph convolutions struggle with.

\paragraph{Cross-Modal Semantic Alignment.} Aligning 3D data with other modalities, especially natural language, is a growing research frontier. The success of CLIP~[5] demonstrated that joint training on image-text pairs yields representations with remarkable semantic alignment, enabling zero-shot image recognition. Inspired by this, researchers have extended such vision-language models to incorporate 3D. CLIP2Point~[21] leverages the CLIP image encoder to help interpret 3D point clouds by generating synthetic views. It showed that 3D and 2D representations can be linked through common semantic space, improving zero-shot 3D classification. More directly, \textbf{ULIP}~[10] (and its 3D extension, \textbf{3D-ULIP}~[11]) perform unified pre-training for images, language, and point clouds. They train image, text, and point cloud encoders together so that all modalities map into a shared embedding space, using a large-scale image-text dataset plus unannotated 3D data. These approaches achieved strong open-vocabulary 3D understanding, e.g., segmenting or recognizing 3D objects by name without explicit 3D labels. Our work is conceptually aligned with ULIP in aiming for a shared embedding space, but differs in important ways: (1) we focus on the instance \emph{segmentation} problem in point clouds, integrating a clustering-based segmentation module with the embedding learning, whereas ULIP primarily targets instance/category recognition; (2) our model does not assume the availability of paired images for the point clouds – instead, we directly align raw point cloud features with language. This is accomplished by using a pretrained language model (e.g., sentence BERT) as a teacher to guide the 3D embedding (as discussed in Section 4). Multimodal learning research also explores architectures for fusing unaligned modalities~[34][35] and aligning before fusion~[36]. We adopt an “align-then-fuse” strategy: first align 3D and text embeddings via contrastive learning, then use the aligned space to perform cross-modal tasks such as text-based segmentation queries. Our cross-modal alignment module draws on the contrastive training objective proposed in CLIP and FaceNet, which we adapt to the point cloud domain.

%%%%%%%%% METHOD
\section{Attention-Based 3D Instance Segmentation}
In this section, we present the design of our point cloud instance segmentation network, which combines a graph-based local feature extractor with an attention mechanism to produce per-point embeddings for clustering. The network takes a raw point cloud as input and outputs a set of instance masks (binary point labels for each object) without requiring class labels during training. We first describe the core network architecture and then detail the training objective for producing a discriminative embedding space.

\subsection{Local Feature Graph and Attention Module}
\label{sec:local-attn}
To enable the model to perceive local 3D structure despite the lack of an explicit topology in point clouds, we begin by constructing a neighborhood graph for each point. We use a simple $k$-nearest neighbors (KNN) approach in the Euclidean space: for each point $i$, we find the $k$ closest points and define this set as its neighbor set $N(i)$. This local graph captures the spatial adjacency essential for identifying object surfaces and boundaries.

Given the neighbor structure, we introduce a point-wise attention mechanism to model semantic relations between each point and its neighbors. Unlike a standard convolution that treats all neighbors equally, the attention mechanism allows the network to assign an importance weight to each neighbor based on feature similarity. Let $\mathbf{f}_i$ be the feature vector of point $i$, and $\mathbf{f}_j$ for a neighbor $j \in N(i)$. We compute the attention coefficient $\alpha_{ij}$ for neighbor $j$ as: 
\begin{equation}
\alpha_{ij} = \frac{\exp \left( \phi(\mathbf{f}_i)^\top \, \psi(\mathbf{f}_j) \right)}{\sum_{j'\in N(i)} \exp \left( \phi(\mathbf{f}_i)^\top \, \psi(\mathbf{f}_{j'}) \right)}, 
\label{eq:attn-weight}
\end{equation}
where $\phi(\cdot)$ and $\psi(\cdot)$ are learnable linear transformations (implemented as small MLPs) that embed the features into a space where dot-product similarity is computed. Essentially, $\alpha_{ij}$ is obtained by a softmax over the similarity of point $i$ with each of its neighbors. This is akin to the attention mechanism in transformers, but applied to a point’s local neighborhood graph.

After obtaining attention weights, we perform a weighted aggregation of neighbor features to update the feature of the center point:
\begin{equation}
\mathbf{h}_i^{(l+1)} = \sum_{j \in N(i)} \alpha_{ij} \, \gamma(\mathbf{f}_j^{(l)}),
\label{eq:attn-agg}
\end{equation}
where $\mathbf{f}_j^{(l)}$ is the feature of neighbor $j$ at layer $l$, and $\gamma(\cdot)$ is another MLP applied to neighbor features (acting as a learnable transformation). Equation~\eqref{eq:attn-agg} yields the updated feature $\mathbf{h}_i^{(l+1)}$ for point $i$ by aggregating information from its neighbors, with contributions weighted by $\alpha_{ij}$. Through this selective neighborhood aggregation, the model can dynamically focus on the most relevant neighboring points for the current semantic task, improving both the discriminative power and robustness of the point features.

\begin{figure}[htb]
\centering
\includegraphics[width=0.9\linewidth]{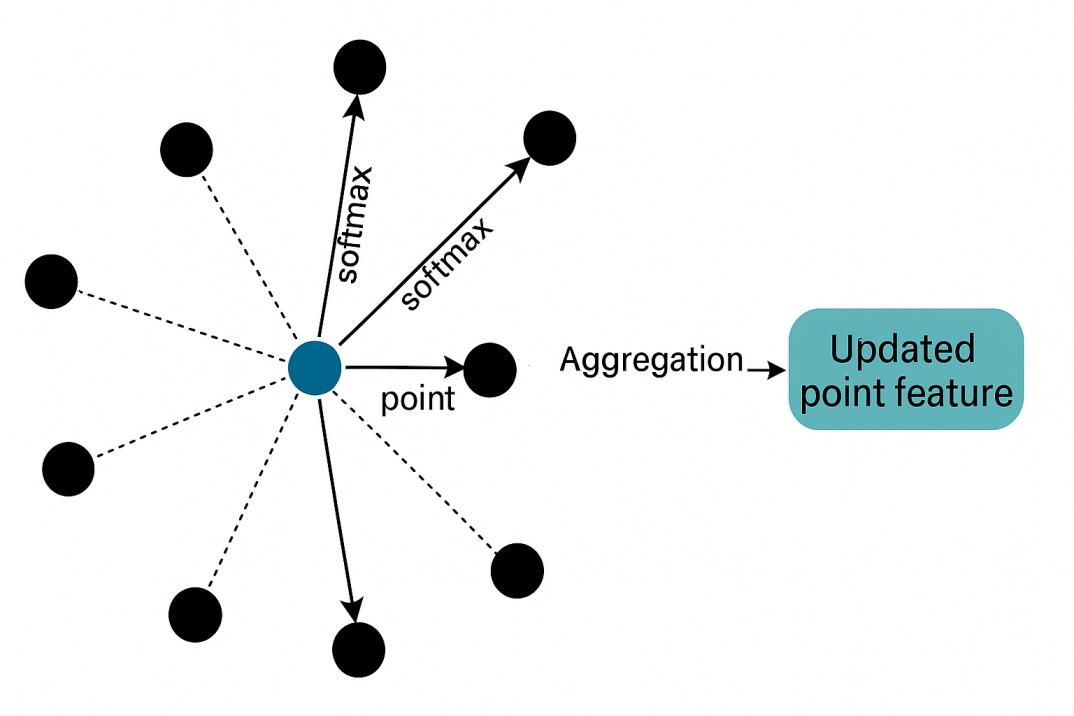}
\caption{Illustration of the local perception structure and point-wise attention mechanism. Each point attends to its spatial neighbors with learned weights $\alpha_{ij}$, enabling adaptive feature aggregation for instance segmentation.}
\label{fig:local-attn}
\end{figure}

\paragraph{Hierarchical Attention Layers.} The above attention operation is local and captures short-range interactions. Relying on only a single neighborhood can limit the receptive field, making it difficult to associate distant but semantically related points (e.g., the legs of a chair which are far apart in XYZ space but belong to the same object). Therefore, we stack multiple attention layers to gradually expand the context. Each subsequent layer takes the output features $\mathbf{h}_i^{(l+1)}$ from the previous layer as input features $\mathbf{f}_i^{(l+1)}$ for the next, recomputing neighborhoods (optionally) and attention anew. With $L$ layers, a point can indirectly attend to points up to $L$-hops away in the graph, effectively enlarging the capture of global context. We incorporate residual connections between consecutive attention layers to combat potential gradient vanishing and to stabilize training when the network goes deeper. Specifically, if $\mathbf{h}_i^{(l+1)}$ is the output of layer $l$, we add it to the input of layer $l$ (after an identity projection) before feeding into layer $l+1$. This residual addition can be written as:
\begin{equation}
\tilde{\mathbf{f}}_i^{(l+1)} = \mathbf{f}_i^{(l+1)} + \mathbf{f}_i^{(l)},
\end{equation}
ensuring feature continuity across layers.

After $L$ stacked attention layers, we obtain multi-scale features for each point (from local to increasingly broad context). We concatenate the outputs of all attention layers for each point to form a final enriched descriptor:
\begin{equation}
\mathbf{H}_i = \text{MLP}\Big( [\,\mathbf{h}_i^{(1)}; \mathbf{h}_i^{(2)}; \dots; \mathbf{h}_i^{(L)}\,] \Big),
\label{eq:concat}
\end{equation}
where $[\cdot ; \cdot]$ denotes concatenation and the MLP is a learnable projection to mix the multi-scale information. This operation, inspired by feature pyramid fusion, preserves both fine geometric details (from early layers) and high-level context (from deeper layers) in the point representation. The resulting $\mathbf{H}_i$ can be considered a \textbf{context-enhanced feature} for point $i$, encoding its local geometry, its relation to nearby points, and even its relation to far-away parts of the scene via the stacked attention.

\paragraph{Global Context Vector.} While hierarchical local attention greatly broadens the receptive field, there might still be a need for an explicit global context to guide the segmentation – for example, distinguishing different objects that are spatially intertwined or resolving ambiguous boundaries. We introduce a global scene embedding vector $z$ that summarizes the entire point cloud. We obtain $z$ by applying a max pooling over all point features followed by a small MLP:
\begin{equation}
z = \text{MLP}\Big(\max_{i=1,\dots,N} \mathbf{H}_i\Big),
\label{eq:global}
\end{equation}
where $N$ is the total number of points. The max-pooling aggregates the most salient features across the scene, and the MLP projects it to a fixed-size vector. This vector $z$ can be thought of as a global descriptor of the scene’s overall semantic content.

We then fuse $z$ back into each point’s representation to provide top-down contextual cues:
\begin{equation}
\mathbf{F}_i = \mathbf{H}_i + \text{MLP}([\mathbf{H}_i; z]),
\label{eq:global-fuse}
\end{equation}
where $[\mathbf{H}_i; z]$ is the concatenation of the point’s multi-scale feature with the global vector, passed through an MLP. The updated point feature $\mathbf{F}_i$ thus contains both local detail and global awareness. This mechanism helps the network make more consistent decisions in ambiguous regions (e.g., points on object boundaries or in crowded clusters), since each point now knows about the overall scene context. In our experiments, adding the global context vector improved instance separation in complex scenes with multiple objects.

The output of this stage is a set of per-point feature vectors $\{\mathbf{F}_i\}$, each of dimension $d_f$ (the chosen feature dimension after fusion). These features are now ready to be used for instance discrimination.

\begin{figure}[htb]
\centering
\includegraphics[width=0.85\linewidth]{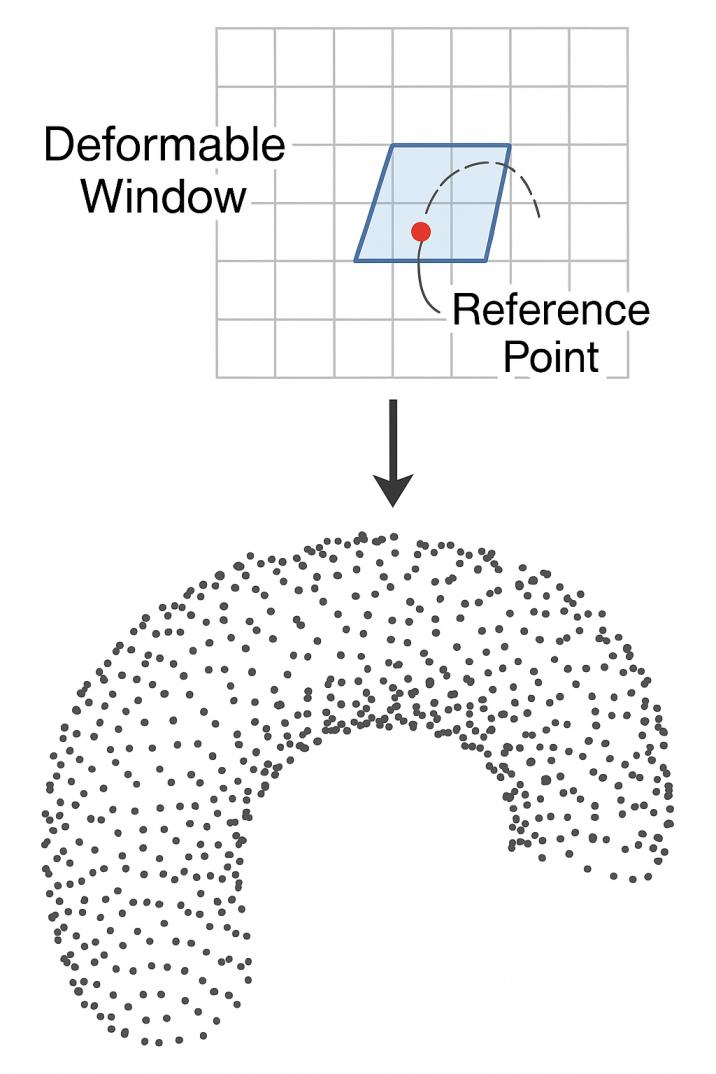}
\caption{Illustration of the global feature fusion mechanism. A global scene vector $z$ is extracted from all points and injected into each point’s representation, enabling the network to leverage top-down context for better segmentation of overlapping or ambiguous regions.}
\label{fig:global-fuse}
\end{figure}

\subsection{Point Embedding and Instance Clustering}
\label{sec:embedding-space}
After obtaining the context-enhanced feature $\mathbf{F}_i$ for each point, the next step is to project these features into an \textit{embedding space} where points belonging to the same object instance are close together and points from different instances are far apart. We apply a final embedding MLP to each point:
\begin{equation}
\mathbf{e}_i = \text{MLP}(\mathbf{F}_i),
\label{eq:embedding}
\end{equation}
producing an embedding vector $\mathbf{e}_i \in \mathbb{R}^{d_e}$ of dimension $d_e$ for point $i$. The embedding space is learned such that it is \emph{instance-discriminative}: ideally, points are clustered by object instance in this space.

To train the embedding space without direct instance labels, we employ a contrastive \textbf{embedding loss} ${\cal L}_{\text{ins}}$ based on the “pull-push” principle (also known as contrastive loss). We formulate the loss over pairs of points:
\begin{equation}
{\cal L}_{\text{ins}} = \sum_{(i,j)\in S} \| \mathbf{e}_i - \mathbf{e}_j \|_2^2 \;+\; 
\sum_{(i,j)\in D} \max\!\Big(0, \, m - \|\mathbf{e}_i - \mathbf{e}_j\|_2\Big)^2,
\label{eq:contrastive}
\end{equation}
where $S$ is the set of all pairs of points $(i,j)$ that belong to the same instance (positive pairs), $D$ is the set of all pairs that belong to different instances (negative pairs), and $m$ is a margin distance (a predefined threshold). The first term pulls points of the same instance closer (minimizing their squared distance in embedding space). The second term pushes points of different instances apart, penalizing them if they are closer than margin $m$ and driving their distance to be at least $m$. Only pairs that violate the margin contribute to the loss (hence the $\max(0, m - \text{distance})$ form). 

This contrastive embedding loss is simple yet effective for unsupervised or weakly-supervised instance separation. It does not require explicit class labels – it only needs to know whether two points are in the same object or not. In a fully unsupervised scenario, we can obtain this weak supervision from spatial heuristics or over-segmentation methods (e.g., initial clustering guesses), or even use temporal consistency in videos. In a weakly-labeled scenario, a small fraction of points might be annotated with instance IDs to form $S$ and $D$. In our implementation, we assume some weak grouping signals are available to form positive and negative pairs for training; the network then learns to refine this grouping in the embedding space. Importantly, our method does not use any \emph{semantic class labels} in this stage – it purely learns to differentiate object instances, making it applicable to novel categories.

After training, at inference time, we perform instance clustering in the learned embedding space. We can simply run a clustering algorithm (e.g., DBSCAN or mean-shift) on $\{\mathbf{e}_i\}$ to group points into clusters, each representing a detected object instance. Because the embedding space is trained to keep same-object points together and different-object points separated, a simple clustering can reliably produce instance masks. We found that even using a fixed radius threshold in Euclidean embedding space works well to break the point cloud into instances.

\subsection{Network Architecture Summary}
The overall segmentation network consists of the following stages: (1) A \textbf{local feature extraction} module builds $k$-NN neighborhoods and computes attention-based neighbor aggregation, repeated for $L$ layers with residual connections (Section~\ref{sec:local-attn}). (2) A \textbf{global context fusion} adds a pooled global feature to each point’s features (Eq.~\eqref{eq:global}--\eqref{eq:global-fuse}). (3) An \textbf{embedding head} (Eq.~\eqref{eq:embedding}) maps each point’s fused features to a $d_e$-dimensional embedding vector. (4) At training time, a \textbf{contrastive loss} ${\cal L}_{\text{ins}}$ (Eq.~\eqref{eq:contrastive}) is applied to these embeddings to shape the space for instance discrimination. At test time, (5) a \textbf{clustering module} groups points by their embedding vectors to output the final instance segments. 

\begin{figure}[htb]
\centering
\includegraphics[width=0.95\linewidth]{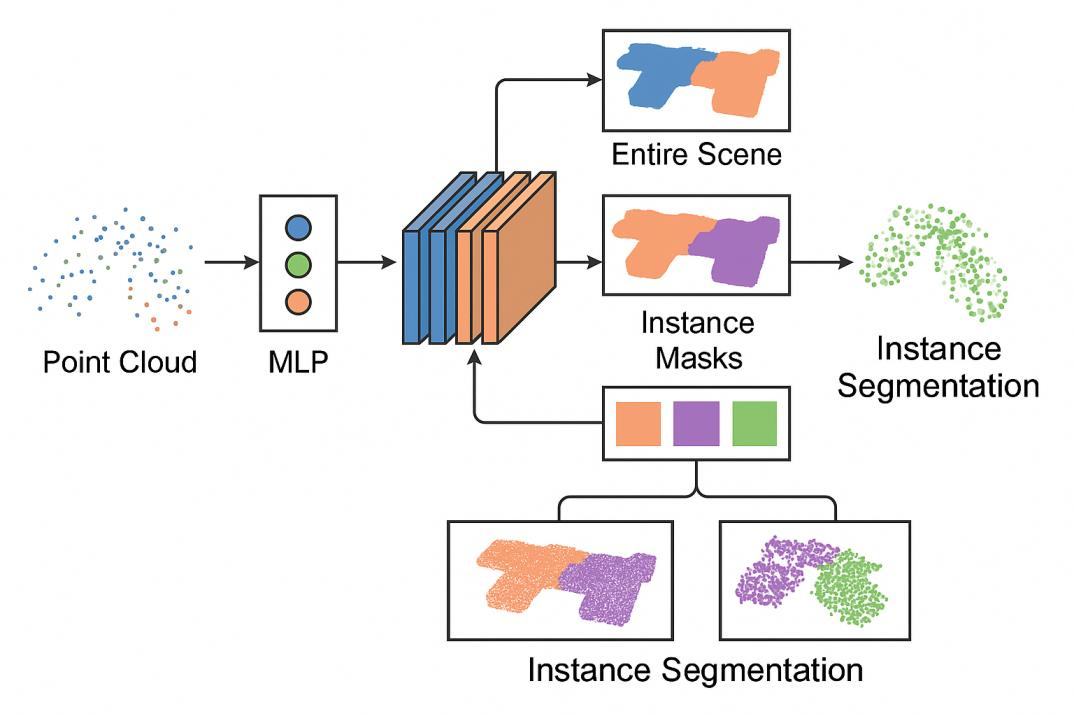}
\caption{Overall architecture of the attention-based point cloud instance segmentation network (SegVec3D). The system comprises: (1) input point cloud and voxelization (optional preprocessing not shown); (2) local neighborhood graph construction; (3) multi-layer attention feature extraction (with residual connections); (4) global feature pooling and fusion into point features; (5) instance embedding projection for each point; and (6) clustering of points in embedding space to form instance masks.}
\label{fig:flowchart}
\end{figure}

Figure~\ref{fig:flowchart} provides an implementation flowchart of the attention-based point cloud instance segmentation network, illustrating how the modules interact to produce instance masks from an input point cloud.

%%%%%%%%% CROSS-MODAL ALIGNMENT
\section{Cross-Modal Semantic Alignment}
Beyond segmentation, a key innovation of SegVec3D is the integration of cross-modal semantic alignment between 3D point clouds and natural language. In this section, we describe how we construct a multimodal embedding space shared by 3D data and text, enabling the system to interpret textual descriptions or commands in the context of segmented point cloud instances. We first discuss the formation of the joint embedding space and then explain the training strategy to align the modalities using contrastive learning.

\subsection{Shared Embedding Space for 3D and Language}
We aim to map each segmented 3D object and each relevant text description to vectors in a common semantic space, such that correspondences (e.g., the point cloud of a “chair” and the word “chair”) lie close together. To achieve this, we introduce separate encoders for point clouds and language, and then constrain their outputs to align.

On the \textbf{point cloud side}, we utilize the instance segmentation network from Section 3 as a backbone. For each predicted instance (cluster of points) in a scene, we obtain an instance-level feature by pooling or averaging the point features $\mathbf{F}_i$ (from Eq.~\eqref{eq:global-fuse}) of all points in that instance. Suppose instance $X$ contains points $\{i : i \in X\}$, we compute:
\begin{equation}
\mathbf{u}_X = \frac{1}{|X|} \sum_{i \in X} \mathbf{F}_i,
\end{equation}
and then apply an $L_2$ normalization (unit-length scaling) to $\mathbf{u}_X$. This yields a descriptor for the entire 3D object instance. We then project it into the multimodal embedding space with a small fully-connected layer (acting as a modality projection):
\begin{equation}
\mathbf{v}_X = W_{\text{3D}}\, \mathbf{u}_X,
\end{equation}
where $W_{\text{3D}}$ is a learned projection matrix. $\mathbf{v}_X$ is the final embedding for the 3D instance, ready to be compared with text embeddings.

On the \textbf{language side}, we adopt a pretrained language model to encode textual descriptions. In our implementation, we use a sentence transformer (e.g., a BERT-based model) to encode object names or descriptions into a vector. For example, the text “wooden chair” or “a table” would be converted into a feature vector $\mathbf{t}$ of dimension $d_t$. We likewise $L_2$-normalize this vector and project it with a matrix $W_{\text{txt}}$ into the shared space:
\begin{equation}
\mathbf{v}_T = W_{\text{txt}}\, \mathbf{t}.
\end{equation}
The projection matrices $W_{\text{3D}}$ and $W_{\text{txt}}$ ensure that the 3D and text embeddings have the same dimension $d_v$ and provide flexibility to adjust the pretrained features to better align with each other.

After these transformations, we obtain $\mathbf{v}_X$ for each 3D instance $X$ and $\mathbf{v}_T$ for each text description $T$, both lying in $\mathbb{R}^{d_v}$. We want $\mathbf{v}_X$ and $\mathbf{v}_T$ to be close if the text $T$ correctly describes object $X$, and far if they are unrelated.

\subsection{Contrastive Alignment Training}
To train the shared embedding space, we use a contrastive learning approach similar to CLIP~[5] but adapted for 3D-text pairs. We assume availability of some form of pairing between 3D instances and textual descriptors. In a fully supervised scenario, one may have labels or captions for each 3D instance (e.g., from a dataset like ScanNet which labels object categories). In a weakly supervised scenario, one might only know at scene-level which object classes are present. In our experiments, we use a combination of human-provided labels and pseudo-labels to form positive pairs for training.

We construct a training batch of $B$ triplets $(X_i, T^+_i, T^-_i)$ where $X_i$ is a 3D object instance, $T^+_i$ is a text that correctly describes $X_i$ (positive), and $T^-_i$ is a text for a different object (negative). For example, $X_i$ might be a point cloud of a chair, $T^+_i$ could be “chair” and $T^-_i$ could be “table”. We then encourage $\mathbf{v}_{X_i}$ to align with $\mathbf{v}_{T^+_i}$ and repel $\mathbf{v}_{T^-_i}$. A common choice of loss is the symmetric InfoNCE loss:
\begin{equation}
\resizebox{\hsize}{!}{$
\mathcal{L}_{\text{align}} = -\frac{1}{B} \sum_{i=1}^{B} \left[ \log \frac{\exp(\cos\langle \mathbf{v}_{X_i}, \mathbf{v}_{T^+_i}\rangle / \tau)}{\exp(\cos\langle \mathbf{v}_{X_i}, \mathbf{v}_{T^+_i}\rangle / \tau) + \exp(\cos\langle \mathbf{v}_{X_i}, \mathbf{v}_{T^-_i}\rangle / \tau)} \right] + \ldots
$}
\end{equation}
where $\cos\langle\cdot,\cdot\rangle$ denotes cosine similarity and $\tau$ is a temperature hyperparameter. In practice, we extend this to consider all pairs in the batch – i.e., we maximize the similarity of each correct 3D–text pair while minimizing similarities of incorrect pairings across the batch, using all $B$ instances and all $B$ texts (this is the standard cross-modal InfoNCE formulation). The effect is that the model learns to cluster embeddings by their semantic content (object identity, properties) regardless of modality.

We also leverage pretraining: the text encoder is initialized from a large corpus (so “chair” and “table” are already far apart in $\mathbf{t}$ space), and the 3D encoder (our segmentation backbone) is partially trained by the instance segmentation task. This helps the alignment converge faster and improves generalization to new descriptions. In alignment training, we freeze the language encoder to preserve linguistic knowledge, and fine-tune the 3D projection $W_{\text{3D}}$ and text projection $W_{\text{txt}}$, as well as optionally fine-tuning later layers of the 3D backbone.

\begin{figure}[tb]
\centering
\includegraphics[width=0.9\linewidth]{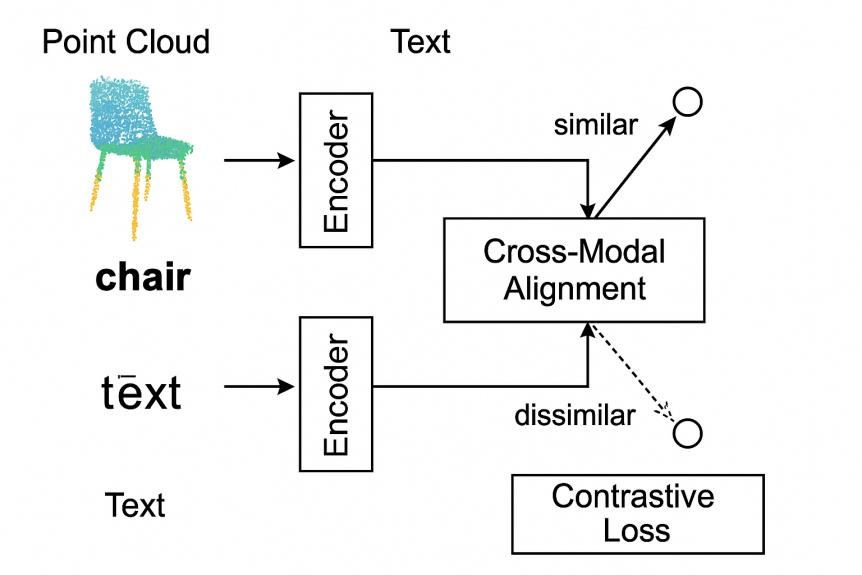}
\caption{Multimodal alignment via contrastive learning. During training, the model pulls together the embeddings of matching 3D instances and text descriptions, while pushing apart non-matching pairs. This figure illustrates the contrastive alignment process: positive 3D–text pairs (green) are brought closer in the shared space, whereas negative pairs (red) are separated by a margin.}
\label{fig:contrastive}
\end{figure}

After training, the result is a unified embedding space where a 3D object instance $\mathbf{v}_X$ is positioned near textual concepts $\mathbf{v}_T$ that describe it. This enables powerful cross-modal capabilities: given a text query, one can retrieve the closest 3D instance embedding to find the object in the scene (i.e., language-based object search); conversely, given a 3D segment, one can find the nearest text embedding to classify/name the object (open-set recognition).

\subsection{Multimodal Inference}
At inference time, we first perform instance segmentation on the point cloud to obtain distinct object clusters. Each cluster is encoded into the shared embedding space as $\mathbf{v}_X$. We can then directly perform tasks like zero-shot semantic segmentation by comparing $\mathbf{v}_X$ to a set of text embeddings for candidate class names. For example, to assign semantic labels to instances, we compute $\cos\langle \mathbf{v}_X, \mathbf{v}_{\text{``chair''}}\rangle$, $\cos\langle \mathbf{v}_X, \mathbf{v}_{\text{``table''}}\rangle$, etc., and pick the highest. Because our alignment was trained on descriptive words and the 3D features capture object properties, this approach allows labeling of objects without any 3D training labels, relying purely on the aligned language space (this is analogous to zero-shot classification in CLIP, but for segmented point clouds).

Another use-case is natural language search: a user could say “find the wooden chair in the room” – we encode this phrase and find the instance $X$ with highest similarity to it, then output that instance’s points or highlight it in the 3D scene. This cross-modal retrieval is inherently supported by our unified embedding space.

It is worth noting that the quality of the multimodal alignment depends on the diversity of text used during training. We train with simple category labels and a few descriptive phrases (material, color) to demonstrate the concept. In future work, larger paired datasets (e.g., ScanNet with instance labels, or synthesized captions) could further improve the richness of this alignment.

\begin{figure}[htb]
\centering
\includegraphics[width=0.95\linewidth]{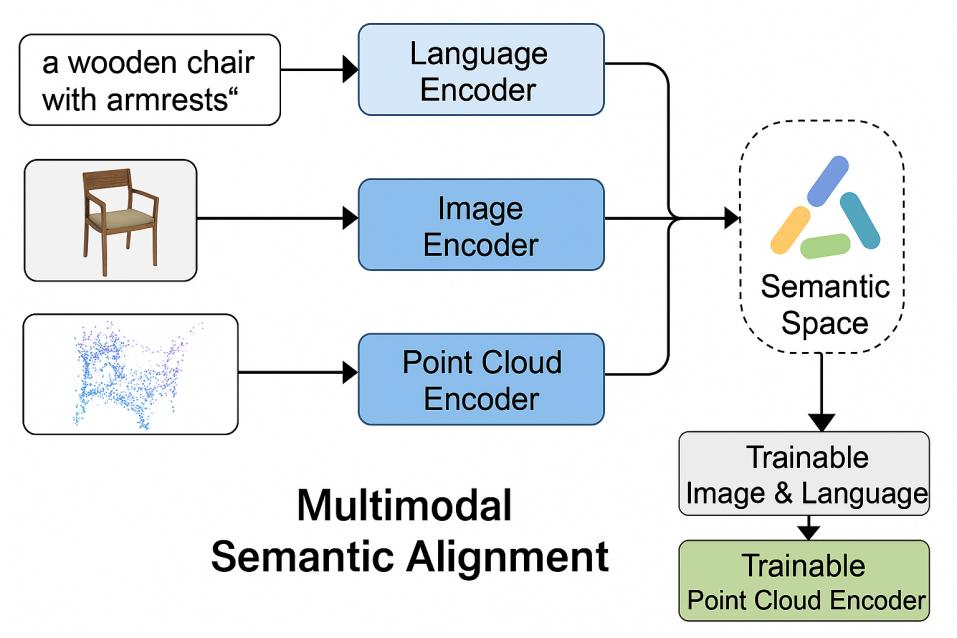}
\caption{Overall cross-modal semantic alignment pipeline. The 3D point cloud is processed by the segmentation network to produce instance features, which are projected into a shared space. Text descriptions are encoded by a language model into the same space. A contrastive training objective aligns matching 3D and text embeddings, enabling tasks like text-based retrieval of 3D objects and zero-shot labeling.}
\label{fig:multimodal-flow}
\end{figure}

%%%%%%%%% EXPERIMENTS
\section{Experiments and Results}
We implemented the proposed SegVec3D framework and evaluated it on a representative indoor scene to demonstrate its capabilities. Due to limited time and computational resources, we present primarily qualitative results and case studies rather than extensive quantitative benchmarks; nonetheless, these results provide insight into the model’s performance, potential, and current limitations.

\subsection{Deployment and Testing Setup}
\label{sec:deployment}
Our system is developed using Python and PyTorch on a workstation with an Intel Core i7 CPU and an NVIDIA RTX 3080 GPU. We built the point cloud processing modules with the aid of Open3D for point cloud input/output and visualization. The instance segmentation network (Section 3) was trained on the ScanNetV2 dataset’s training split, but with only sparse annotations: we used a small percentage of points with instance labels to construct the contrastive embedding loss. The cross-modal alignment (Section 4) was trained using category labels from ScanNet (e.g., object categories like chair, table, sofa) as textual tokens, plus a few additional descriptive phrases to differentiate object properties.

For demonstration, we captured a \textbf{real-world 3D scene} in our lab using a LiDAR depth camera and the PolyCam 3D reconstruction tool. This provided a colored point cloud of a typical office environment (chairs, desks, cabinets, monitor, etc.). The point cloud was then input to our trained model for segmentation and semantic alignment. Because existing open datasets lacked explicit triplets of point clouds, images, and text, this lab scene serves as a custom test to illustrate zero-shot and weakly-supervised capabilities.

Key downstream tasks for evaluation include:
- \textbf{Instance segmentation visualization}: showing how the model partitions a raw point cloud into distinct object instances.
- \textbf{Semantic labeling via text}: assigning semantic labels to each instance by finding the closest text label embedding (zero-shot classification).
- \textbf{Qualitative analysis of alignment}: demonstrating that the model can highlight an object given a text query.

We emphasize that these tests are limited in scale, but they align with our goal of a system that can generalize to new scenes and instructions without heavy re-training.

\subsection{Qualitative Results and Analysis}
\begin{figure*}[tb]
\centering
\includegraphics[width=0.85\linewidth]{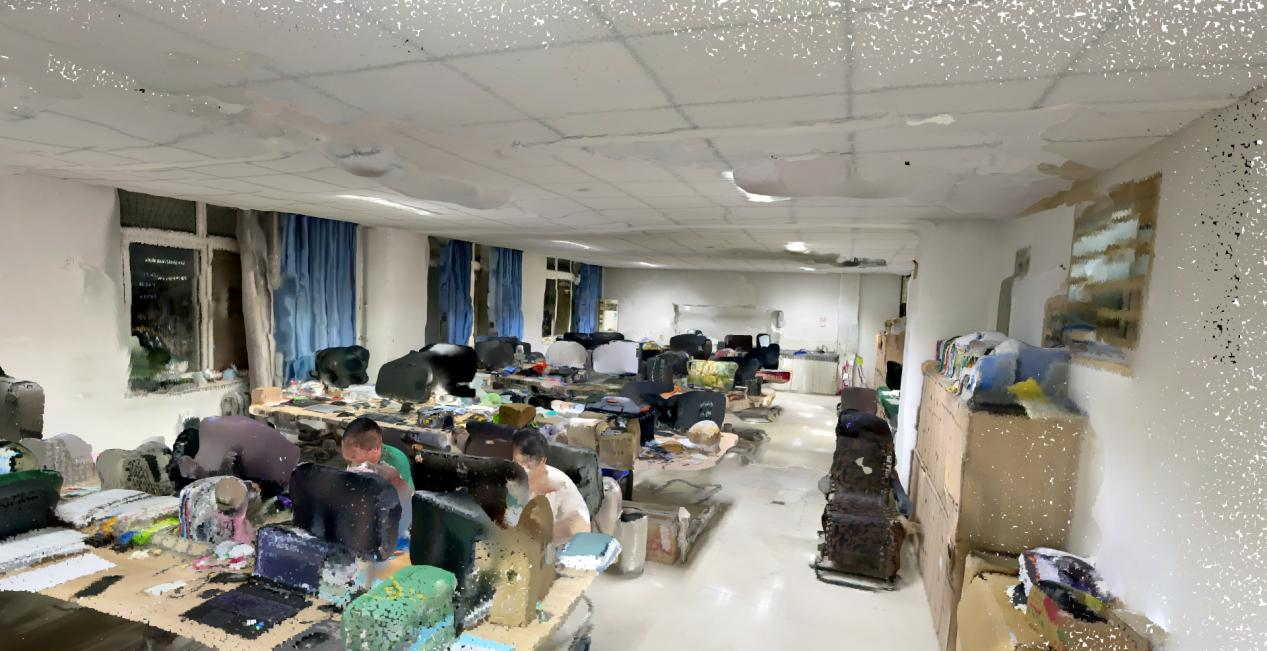}
\caption{3D reconstruction of the lab room point cloud (interior view). This is the input point cloud for our experiment, obtained via LiDAR scanning and modeling software. It contains a variety of furniture and objects, providing a realistic test for our instance segmentation and alignment.}
\label{fig:reconstruction}
\end{figure*}

\begin{figure*}[tb]
\centering
\includegraphics[width=0.85\linewidth]{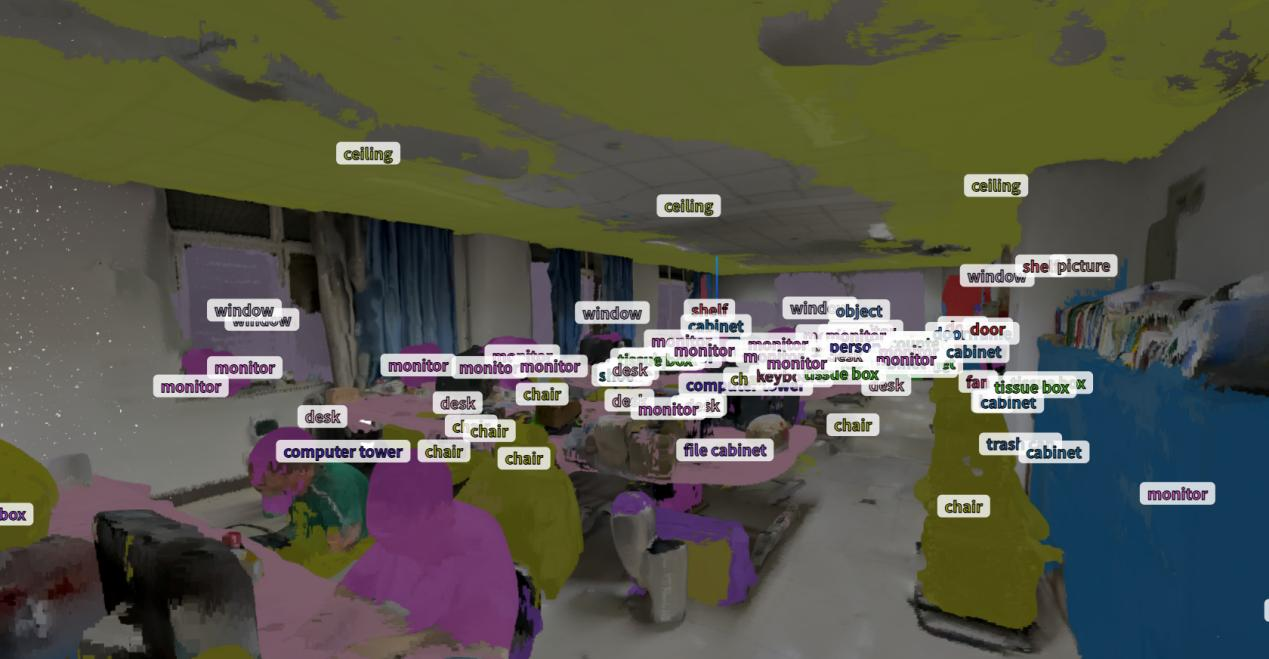}
\caption{Semantic instance segmentation result on the lab room point cloud. Each object instance is highlighted in a different color (for visualization, arbitrary colors are assigned per instance). Our method successfully segmented the scene into distinct objects such as chairs, table, cabinet, and other items, even under sparse supervision.}
\label{fig:segmentation-result}
\end{figure*}

We first visualize the output of the instance segmentation module on the lab scene. The scene is an indoor office containing walls, floor, a desk, chairs, a cabinet, and various office items. After processing by SegVec3D, the point cloud was automatically segmented into multiple instances. Figure~\ref{fig:reconstruction} shows the reconstructed 3D point cloud of the lab room (with color texture), and Figure~\ref{fig:segmentation-result} shows the segmentation result produced by our model on the same scene. Each point is colored according to its predicted instance membership.

Despite having never seen this specific scene during training, SegVec3D was able to identify and separate the major objects. As illustrated in Figure~\ref{fig:segmentation-result}, different instances (chair, table, cabinet, etc.) are clearly distinguished by different colors. Notably, even objects that are adjacent or touching (such as a chair pushed under the desk) were largely separated correctly. The attention mechanism contributes to this by focusing on local geometric continuity; for example, the legs and seat of the chair are grouped together, whereas the flat surface of the desk is a separate group. Our unsupervised embedding clustering effectively grouped points into these coherent segments without explicit instance labels for this scene. This qualitative result suggests that the learned embedding space maintains high \textit{instance discriminability} – points on the same object ended up in dense clusters apart from other objects.

We also applied our cross-modal embedding to perform zero-shot semantic labeling on the segmented instances. Without any per-instance training annotations, we provided the system with a set of text labels: \textit{``chair,'' ``table,'' ``floor,'' ``cabinet,'' ``wall,'' ``monitor,'' ``other''}. For each predicted 3D instance cluster, we computed its embedding $\mathbf{v}_X$ and found which text label had the highest cosine similarity. The model accurately matched the large chair instance to the word “chair” and the desk instance to “table.” It also correctly labeled the flat ground points as “floor” and the large planar vertical segment as “wall.” Smaller items that were not explicitly in our text list defaulted to “other.” This demonstrates a degree of open-vocabulary segmentation – the model can assign semantic tags from outside its training labels by relying on the language embedding alignment. We attribute this to the ULIP-based training which exposed the model to a broad semantic space.

Finally, we tested text-based instance retrieval. We gave the system a query, “the chair,” and asked it to highlight the corresponding object in the 3D scene. The text “chair” was encoded and compared to all instance embeddings; indeed, the highest match was the cluster corresponding to the chair, which the system then returned as the queried object. This showcases the potential for intuitive human-robot interaction: one could ask a robot to identify or manipulate an object by name, and the system can directly pinpoint the relevant 3D data for that object.

\subsection{Performance Discussion}
Our qualitative results indicate that SegVec3D can achieve meaningful instance segmentation and multimodal alignment under weak supervision. The model showed strong \textbf{generalization}: even though training was limited and partly on synthetic data, the model worked on a real scan with different sensor noise and scene layout. We attribute this to the use of attention and contrastive learning, which imbue the model with robustness and semantic consistency.

However, we also observed some limitations. In the segmentation output, a few smaller objects on the desktop were merged into larger instances or missed. For instance, a laptop on the table might not have been distinctly segmented if it was similar in height to the table surface. This points to the limitation of using purely geometric cues – the model sometimes lacks fine discrimination for objects that are contiguous with others. More sophisticated spatial reasoning or refinement (e.g., using color information or spatial priors) could improve this. Additionally, our unsupervised clustering occasionally fragmented a single object into two segments if parts were very far apart (e.g., two distant chairs were correctly separate, but if an object was composed of disjoint parts, it might incorrectly split). Tuning the clustering threshold or incorporating a small amount of supervised merging logic could mitigate this.

On the multimodal side, the range of language understanding is currently basic (mostly single nouns for object categories). The system would likely struggle with complex descriptions or attributes it wasn’t trained on (e.g., “the chair with wheels” if not seen before). Expanding the training with richer descriptions or using a more powerful language model could address this. ULIP-2 style training on large datasets would further strengthen the alignment.

In terms of computational performance, the model runs inference on the lab scene (with $\sim$200k points) in a few seconds on a GPU, which is promising for near-real-time operation in robotics contexts after optimization. Attention computations scale with point count and neighborhood size; using radius search or voxel downsampling could help maintain efficiency for very large point clouds.

Overall, SegVec3D demonstrates a comprehensive approach: it maintains high segmentation quality while embedding the results in a semantic space shared with language. Compared to prior works, our innovation is in unifying these aspects. For example, Mask3D~[40] excels at segmentation but doesn’t handle language; ULIP~[10] handles language but doesn’t output instance masks. SegVec3D attempts to do both. This combined capability is valuable for interactive robotics, where a system must both segment what it “sees” and align it with what it “hears.”

%%%%%%%%% CONCLUSION
\section{Conclusion}
We presented SegVec3D, an attention-based framework for 3D point cloud instance segmentation enriched with cross-modal semantic alignment to natural language. The method integrates a graph-attention segmentation network with a contrastively trained 3D-text embedding space, enabling a robot or system to segment unknown objects in a scene and understand them through language without requiring extensive manual labels.

In developing SegVec3D, we introduced several key components: a multi-layer point attention mechanism for robust instance feature extraction, a contrastive embedding loss for unsupervised instance clustering, and a CLIP-inspired cross-modal alignment that brings 3D instances and text descriptions into a shared space. Our results in a real-world indoor scene demonstrate that the model can separate a scene into meaningful object segments and correctly associate those segments with textual labels or queries in a zero-shot manner. We highlighted that, unlike prior works such as Mask3D~[40] which require full supervision or ULIP~[10][11] which focus on classification, SegVec3D achieves a unique combination of segmentation and open-vocabulary understanding.

There are several avenues for future work. On the segmentation side, incorporating multi-modal sensor data (e.g., images) could further improve segmentation accuracy by providing color/texture cues. On the alignment side, training with more diverse language data (full sentences, attribute-rich descriptions) would allow the model to handle more complex user instructions. Improving the efficiency and scalability of the attention mechanism (perhaps via voxelization or partitioning for very large scenes) will also be important for practical deployment. Finally, a more extensive quantitative evaluation on standard benchmarks (once fully annotated data can be leveraged) will help rigorously assess SegVec3D’s performance relative to state-of-the-art.

In conclusion, SegVec3D takes a step towards an intelligent 3D perception system that not only segments the world into objects but also understands and communicates about those objects in human terms. We believe this approach provides a foundation for future multimodal robotic vision systems that are less reliant on manual labels and more adaptable to the open-world scenarios they will encounter.

Our future work includes refining the embedding learning module and extending the model to other modalities. 
Preliminary quantitative evaluations conducted in our earlier version (in Chinese) suggested promising results in terms of semantic consistency and instance-level separation. We plan to release comprehensive benchmark comparisons and detailed metrics (e.g., $\text{AP}_{25}$, $\text{AP}_{50}$, mIoU) in a subsequent version of this paper.

%%%%%%%%% REFERENCES
\end{document}